\documentclass[letterpaper]{article} 
\usepackage[draft]{aaai2026}  
\usepackage{times}  
\usepackage{helvet}  
\usepackage{courier}  
\usepackage[hyphens]{url}  
\usepackage{graphicx} 
\usepackage{natbib}  
\usepackage{caption} 
\usepackage{amssymb}

\usepackage{algorithm}
\usepackage{algorithmic}
\usepackage{amsmath}
\usepackage{newfloat}
\usepackage{listings}
\usepackage{pgfplots}
\usepackage{pgfplotstable}
\usepackage{caption}
\pgfplotsset{compat=1.17}

\DeclareCaptionStyle{ruled}{labelfont=normalfont,labelsep=colon,strut=off} 
\floatstyle{ruled}
\newfloat{listing}{tb}{lst}{}
\floatname{listing}{Listing}
\title{VGGT-DP: Generalizable Robot Control via Vision Foundation Models}
\author{
    Shijia Ge\textsuperscript{\rm 1},
    Yijun Liu\textsuperscript{\rm 1},
    Yinxin Zhang\textsuperscript{\rm 2},
    Shuzhao Xie\textsuperscript{\rm 1},
    Weixiang Zhang\textsuperscript{\rm 1},
    Mingcai Zhou\textsuperscript{\rm 3},
    Zhi Wang\textsuperscript{\rm 1}
}
\affiliations{
    \textsuperscript{\rm 1}Shenzhen International Graduate School, Tsinghua University, Shenzhen, China \\
    \textsuperscript{\rm 2}Harbin Institute of Technology, Shenzhen, China \\
    \textsuperscript{\rm 3}CASBOT, Beijing, China
}

\usepackage{bibentry}

\begin{document}

\maketitle
\begin{abstract}
Visual imitation learning frameworks allow robots to learn manipulation skills from expert demonstrations. While existing approaches mainly focus on policy design, they often neglect the structure and capacity of visual encoders—limiting spatial understanding and generalization. Inspired by biological vision systems, which rely on both visual and proprioceptive cues for robust control, we propose \emph{VGGT-DP}, a visuomotor policy framework that integrates geometric priors from a pretrained 3D perception model with proprioceptive feedback. We adopt the Visual Geometry Grounded Transformer (VGGT) as the visual encoder and introduce a \emph{proprioception-guided visual learning} strategy to align perception with internal robot states, improving spatial grounding and closed-loop control. To reduce inference latency, we design a \emph{frame-wise token reuse (FTR) mechanism} that reuses cached VGGT aggregator tokens from overlapping observation frames and computes features only for the latest frame, substantially reducing redundant visual encoding. We further apply \emph{random token pruning} to enhance policy robustness and reduce overfitting. Experiments on challenging MetaWorld tasks show that VGGT-DP significantly outperforms strong baselines such as DP and DP3, particularly in precision-critical and long-horizon scenarios.
\end{abstract}

\section{Introduction}

\begin{figure}[H]
    \centering
    \includegraphics[width=0.9\linewidth]{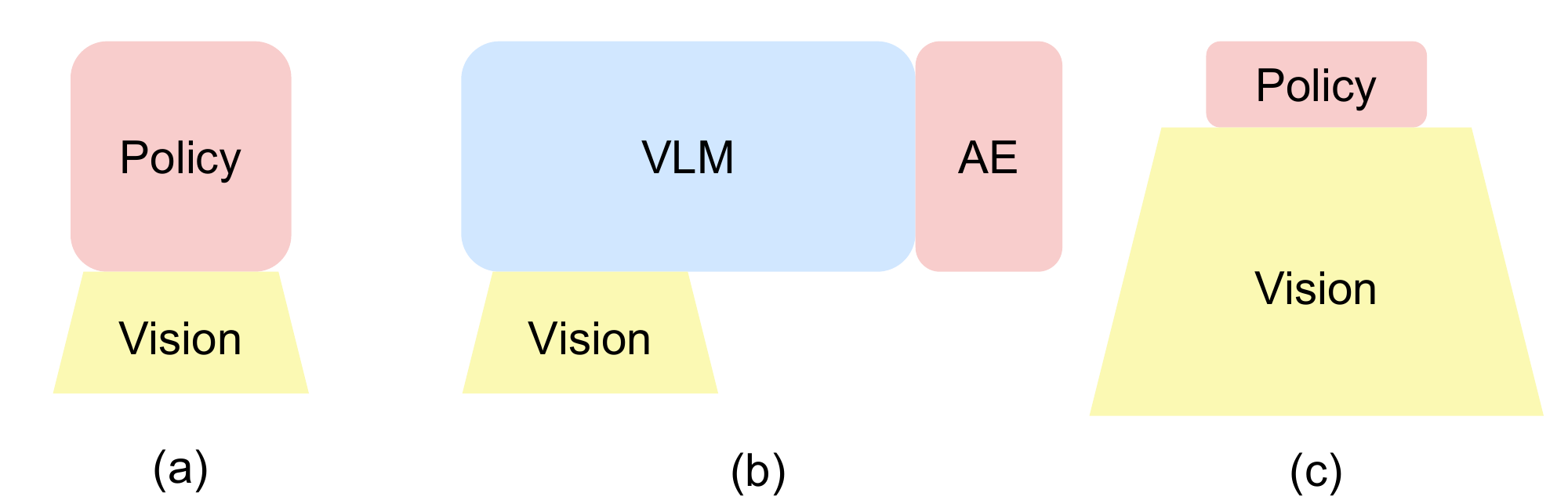}
    \caption{Comparison of current paradigms for visual robot policies. (a) Single-task Vision-Action (VA) paradigm with a relatively small vision encoder and a large policy head. (b) Vision-Language-Action (VLA) paradigm, typically employing a large pretrained vision-language model (VLM) as the backbone paired with a small action generation expert (AE), with the vision encoder slightly larger. (c) Our proposed VGGT-DP paradigm, which utilizes a large-scale pretrained visual encoder and a relatively compact policy head, emphasizes robust spatial representation to facilitate generalizable and stable robotic control.}
    \label{fig:vmp_paradiam}
\end{figure}

Visuomotor robot policies fundamentally rely on sensory inputs to infer the robot's current state and understand its environmental context, enabling tasks such as grasping, manipulation, and navigation. Historically, robotic systems were explicitly programmed using handcrafted rules and analytical models, which significantly limited their flexibility and adaptability in dynamic environments. The past decade has witnessed a transformative shift toward data-driven approaches, wherein robots directly learn control policies from large-scale, diverse datasets containing sensor observations paired with corresponding actions. Typically, these sensory modalities encompass visual inputs, proprioceptive signals from the robot's internal sensors, and often natural language instructions. Fig.~\ref{fig:vmp_paradiam} illustrates the currently prevalent paradigms in visuomotor robot policy research.

Recent robotic research has increasingly emphasized incorporating large language models (LLMs) into robotic control frameworks, known as vision-language-action (VLA) models~\cite{brohan2022rt1,zitkovich2023rt2,pi0,kim2024openvla}, driven by the belief that linguistic priors embedded in these models enhance generalization capabilities across various tasks and environmental settings. However, it is noteworthy that advanced manipulation and locomotion capabilities are prominently observed in non-linguistic biological organisms. For instance, animals such as insects, including flies, and even single-celled organisms like paramecia, exhibit remarkable abilities to perceive, navigate, and manipulate their environments without any reliance on linguistic or symbolic reasoning. This observation indicates that sophisticated perception and manipulation skills might depend more fundamentally on rich and extensive visual and proprioceptive sensory processing rather than linguistic abstractions. Motivated by this insight, we propose a language-free visuomotor policy framework that emphasizes geometry-aware vision and proprioceptive feedback, rather than linguistic conditioning. By forgoing language priors, our approach focuses on learning spatially grounded representations and low-level control behaviors that align more closely with the mechanisms observed in biological systems.

Notably, biological studies consistently demonstrate that a significant portion of neural resources in animals is dedicated specifically to visual processing \cite{fruitflynature}. Such an allocation facilitates robust and adaptive interactions with dynamically changing environments, underpinning remarkable task generalization. Conversely, the visual encoders commonly used in current robotic systems prioritize computational efficiency, often resulting in overly simplistic and low-capacity architectures that struggle to capture complex spatial and geometric relationships within visual scenes. Consequently, even small shifts in the environment can lead to significant degradation in robotic manipulation performance due to insufficient representation capacity.

 Inspired by this observation, we seek to endow robotic systems with analogous capabilities by leveraging large-scale pretrained vision models such as VGGT. These models, trained on extensive 3D reconstruction tasks, can serve as powerful visual priors to enable spatially grounded and semantically meaningful representations—thus simulating the perception-driven control behavior observed in nature. Our work integrates the VGGT into a Diffusion Policy (DP) framework. By combining the robust geometric grounding provided by VGGT with the probabilistic reasoning capabilities of DP, our framework aims to significantly enhance the robustness and generalization capabilities of vision-based robotic manipulation.

In summary, our work bridges biological inspiration and contemporary robotic control methodologies, highlighting the critical role of large-scale visual representation in achieving robust robot manipulation. This paper makes the following contributions:
\begin{itemize}
    \item We propose \textbf{VGGT-DP}, a visuomotor control framework that integrates a 3D-reconstruction-pretrained visual foundation model (VGGT) with diffusion policy, enhancing spatial perception and action planning in manipulation tasks.
    \item We design a \textbf{Frame-Wise Token Reuse (FTR) mechanism} and a \textbf{Random Token Pruning strategy} to reduce computational overhead while improving the robustness of visual representations.
    \item We introduce a \textbf{proprioception-guided visual learning method} to better align visual features with internal robot states, thereby improving closed-loop feedback control.
    \item We empirically validate the framework on diverse challenging MetaWorld tasks, showing that VGGT-DP outperforms strong baselines in precision-demanding scenarios.
\end{itemize}
\section{Preliminary}

\subsection{Diffusion Policy}
\label{sec:prelim_diffusion}

Diffusion policies ~\cite{chi2023diffusionpolicy}offer a novel and effective framework for generating robot actions, enabling them to adapt to dynamic environments and achieve long-term task goals. By modeling the action generation process as a denoising diffusion process, diffusion policies ensure better handling of uncertainty and improve control performance over time.

Diffusion policy models the robot policy as a denoising diffusion process conditioned on past observations. At each time step \(t\), the policy receives the observation sequence:
\[
\mathcal{O}_t = \{o_{t - T_o + 1}, \dots, o_t\},
\]
and predicts a future action sequence:
\[
\hat{A}_t = \{a_t, \dots, a_{t + T_p - 1}\}.
\]

This prediction is generated by iteratively refining an initial noise sample using the learned denoising function \(\epsilon_\theta\). At each diffusion step \(k\), the update rule is:
\[
    A_t^{(k-1)} = \alpha_k \left(A_t^{(k)} - \gamma_k \, \epsilon_\theta(\mathcal{O}_t, A_t^{(k)}, k)\right) + \mathcal{N}(0, \sigma_k^2 I).\]

During execution, only the first \(T_a\) actions are applied:
\[
    \text{Execute } a_t, \dots, a_{t + T_a - 1}; \quad \text{Replan at } t + T_a.\]

This receding-horizon approach maintains long-term temporal consistency while enabling adaptive control in dynamic settings.

\subsection{Visual Geometry Grounded Transformer}

Visual Geometry Grounded Transformer (VGGT) is a unified perception model for extracting 3D-aware visual representations from multi-view images. Given a set of images \(\{I_i\}_{i=1}^N\), this model predicts the canonical geometric outputs:
\[
    \operatorname{VGGT}(\{I_i\}_{i=1}^N) = \left(\{G_i\}_{i=1}^N, \{D_i\}_{i=1}^N, \{P_i\}_{i=1}^N, \mathcal{T}\right),
\]
where \(G_i \in \mathbb{R}^9\) denotes the camera parameters for view \(i\), including rotation \(q \in \mathbb{R}^4\), translation \(t \in \mathbb{R}^3\), and field of view \(f \in \mathbb{R}^2\); \(D_i \in \mathbb{R}^{H \times W}\) is a dense depth map; \(P_i \in \mathbb{R}^{3 \times H \times W}\) is a 3D point map in world coordinates; and \(\mathcal{T}\) denotes 3D point tracks across the input views and frames. These canonical predictions should not be confused with the aggregator tokens used as policy features below.

Internally, VGGT integrates two key components: an aggregator, which fuses spatial and appearance cues from all input images through a multi-layer transformer to produce geometry-aware tokens with global context; and the dense prediction transformer~\cite{dpt}, which generates high-quality depth maps and point cloud representations.

\section{Method}

\subsection{Problem Formulation}
\label{sec:problem_formulation}

We consider the problem of learning a visuomotor policy from expert demonstrations in a robotic manipulation setting. Specifically, the agent is provided with a dataset of expert trajectories $\mathcal{D} = \{(\mathbf{o}_t, \mathbf{a}_t)\}_{t=1}^T$, where $\mathbf{o}_t$ denotes the observation at time $t$, and $\mathbf{a}_t$ is the corresponding expert action. The goal is to learn a policy $\pi: \mathcal{O} \rightarrow \mathcal{A}$ that maps observations to actions and generalizes to unseen states and environments.

We model the robotic manipulation environment as a partially observable Markov decision process (POMDP), defined by the tuple $(\mathcal{S}, \mathcal{A}, \mathcal{O}, P, r, \gamma)$. The state space $\mathcal{S}$ comprises the robot's proprioceptive state (e.g., joint positions and end-effector pose) and the external environment configuration (e.g., object positions). The observation space $\mathcal{O}$ includes both visual inputs $\mathbf{I}_t \in \mathbb{R}^{C \times H \times W}$ and proprioceptive signals $\mathbf{s}_t \in \mathbb{R}^{d_s}$. The action space $\mathcal{A}$ consists of continuous control commands, including Cartesian displacements $\Delta \mathbf{x}_t \in \mathbb{R}^3$ and gripper states $g_t \in \mathbb{R}$, i.e., $\mathbf{a}_t = [\Delta \mathbf{x}_t, g_t]$.

The agent is trained to imitate the expert policy $\pi_E$ using diffusion-based imitation learning, a method that enables robust policy learning despite the partial and potentially noisy nature of observations. At deployment, the learned policy must effectively infer the appropriate actions from partial observations, even when confronted with unseen viewpoints or environmental variations. Our objective is to build a policy that not only mimics expert behavior but also demonstrates resilience to spatial ambiguities by effectively utilizing both visual and proprioceptive cues.

\begin{figure*}[ht]
    \centering
    \includegraphics[width=1\linewidth]{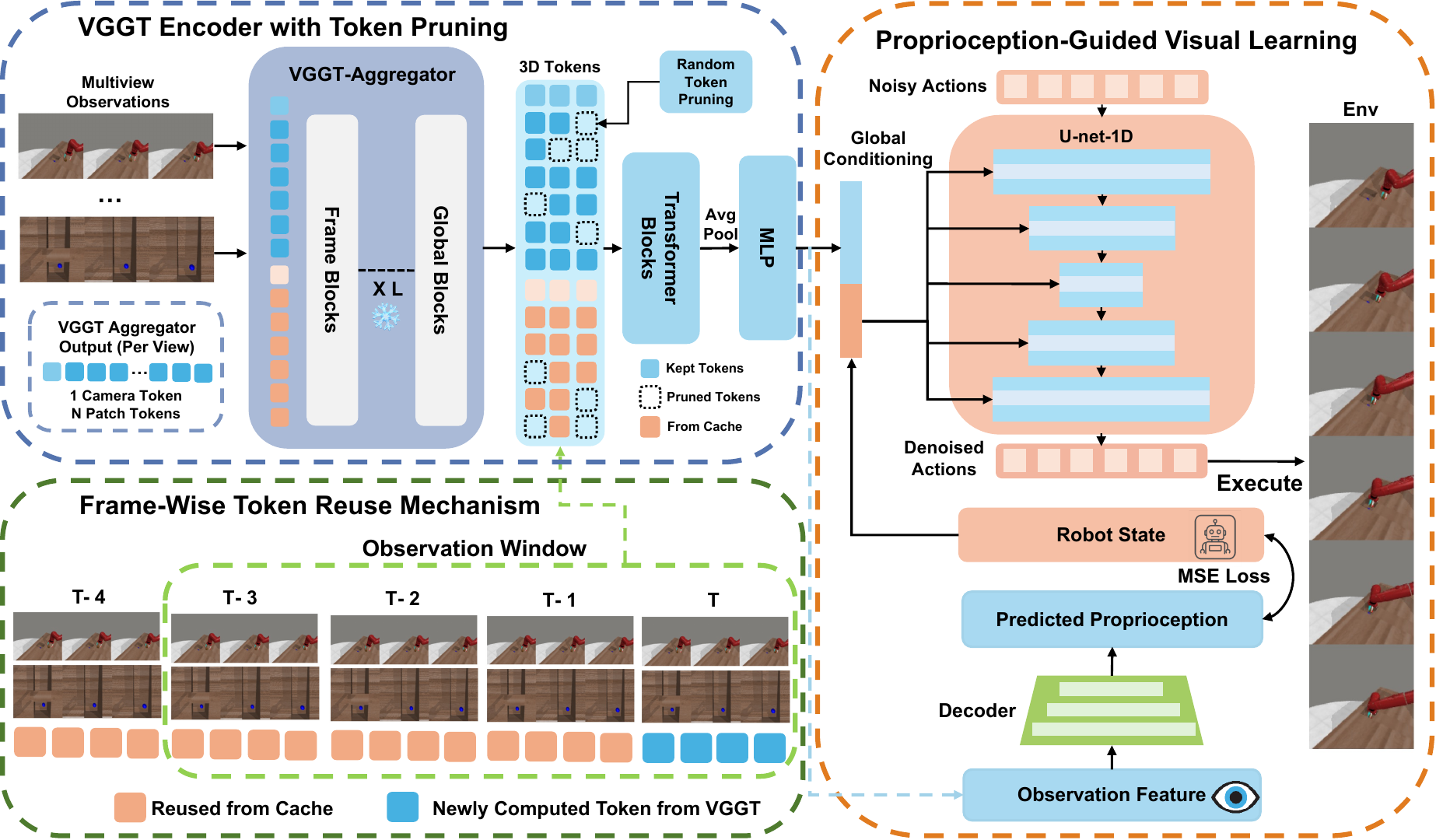}
    \caption{\textbf{Architecture of VGGT-DP}. The system consists of three components: 
  (1) a VGGT encoder with token pruning for extracting geometry-aware features, 
  (2) a frame-wise token reuse mechanism for efficient inference, and 
  (3) a proprioception-guided diffusion policy for multi-step action prediction.}
    \label{fig:placeholder}
\end{figure*}

\subsection{VGGT as the Feature Projector}
\label{sec:vggt_projector}

To achieve our goal, we leverage VGGT as a feature projector for extracting geometry-aware visual representations. Rather than relying on low-level visual outputs, we focus on the aggregated tokens generated by the VGGT aggregator, which serve as compact, semantically rich representations of the 3D scene. These tokens are crucial for grounding visuomotor control and ensuring that the policy can effectively interpret spatial relationships in the environment. The VGGT aggregator is kept frozen during policy training; the downstream Transformer encoder and MLP constitute the trainable visual projector.

Specifically, we use the aggregator output from the pretrained VGGT model. Given a visual observation sequence of \(T\) frames from \(V\) camera views, we input a tensor of shape
\(
\mathcal{I} \in \mathbb{R}^{B\cdot T\times V \times 3 \times H \times W},
\)
where \(B\) is the input batch size, \(T\) is the observation steps, and V is the number of camera views.

VGGT produces the corresponding visual tokens:
\[
\mathcal{T}_{\text{vggt}} = \text{VGGT}_{\text{agg}}(\mathcal{I}) \in \mathbb{R}^{B \cdot T \times V \times (N_p+1) \times D},
\]
where \(N_p\) is the number of patch tokens per view,\(+1\) is the camera token, and \(D\) is the token embedding dimension.

For notational clarity, let \(\mathcal{T}_{\text{in}}\) denote the token tensor provided to the Transformer encoder, and let \(N_{\text{in}}\) be its number of tokens per view. Without pruning, \(\mathcal{T}_{\text{in}}=\mathcal{T}_{\text{vggt}}\) and \(N_{\text{in}}=N_p+1\); with pruning, \(\mathcal{T}_{\text{in}}=\mathcal{T}_{\text{pruned}}\) and \(N_{\text{in}}=(1-r_{\text{prune}})N_p+1\), as defined below. We reshape the tokens into a sequence and pass them through a Transformer encoder:
\[
\mathcal{T}_{\text{proj}} = \text{TransformerEncoder}(\mathcal{T}_{\text{in}}) \in \mathbb{R}^{B \cdot T \times V \cdot N_{\text{in}} \times D}.
\]
We then apply average pooling over the token dimension:
\[
\bar{\mathcal{T}} = \frac{1}{V \cdot N_{\text{in}}} \sum_{i=1}^{V \cdot N_{\text{in}}} \mathcal{T}_{\text{proj}}[:, i, :] \in \mathbb{R}^{B \cdot T \times D}.
\]
To obtain a compact condition embedding suitable for the diffusion model, we use an MLP to project the pooled feature to a low-dimensional condition embedding:
\[
\mathcal{C} = \text{MLP}(\bar{\mathcal{T}}) \in \mathbb{R}^{B \cdot T \times d_c},
\]
where \(d_c\) is the target condition dimension for the Diffusion Policy.

\subsection{Frame-Wise Token Reuse}
\label{sec:ftr}

Existing approaches typically recompute visual embeddings for all observation frames at every inference step, even when frames overlap across time. While this is acceptable for small encoders, it results in significant and unnecessary latency when employing high-capacity models like VGGT. We thus propose a Frame-Wise Token Reuse mechanism to avoid redundant computation and accelerate inference. For observations at time step \(t\), we reuse the tokens from previous frames \(\{t - T + 1, \dots, t-1\}\), which remain unchanged across overlapping observation windows. 

Let \(\mathcal{T}_{\text{cache}}^{(t-1)} \in \mathbb{R}^{B \cdot (T - 1) \times V \times (N_p+1) \times D}\) denote the cached VGGT aggregator tokens for the previous \(T-1\) frames. The cache therefore stores tokens, not image inputs; it may be offloaded to the CPU between policy calls. At time \(t\), we run the frozen VGGT aggregator only on the latest image \(\mathcal{I}_t\) and reconstruct the full token sequence as:
\[
\begin{aligned}
\mathcal{T}_{\text{vggt}}^{(t)} &= \text{Concat}\left(\mathcal{T}_{\text{cache}}^{(t-1)}, \, \text{VGGT}_{\text{agg}}(\mathcal{I}_{t})\right),\\[-1pt]
\mathcal{T}_{\text{vggt}}^{(t)} &\in \mathbb{R}^{B \cdot T \times V \times (N_p+1) \times D}.
\end{aligned}
\]

\subsection{Augmentation via Random Token Pruning}
\label{sec:token_pruning}

To improve representation robustness and prevent overfitting to specific token positions, we apply random token pruning before feeding VGGT tokens into the Transformer encoder. We randomly drop a proportion \(r_{\text{prune}}\) of the \(N_p\) patch tokens for each view, while retaining the camera token:
\[
\begin{aligned}
\mathcal{T}_{\text{pruned}} &= \text{Prune}(\mathcal{T}_{\text{vggt}}, r_{\text{prune}}),\\[-1pt]
\mathcal{T}_{\text{pruned}} &\in \mathbb{R}^{B \cdot T \times V \times \left((1-r_{\text{prune}})N_p+1\right) \times D}.
\end{aligned}
\]

The camera token is never dropped and remains in the Transformer input and subsequent average pooling. Thus the pruned sequence uses \(N_{\text{in}}=(1-r_{\text{prune}})N_p+1\) in the equations above. This strategy introduces stochasticity at the token level, encouraging the model to learn representations that are invariant to partial observation dropout. Also, token pruning~\cite{li-1} serves as a lightweight strategy to enhance inference speed by eliminating redundant patch tokens.

\subsection{Proprioception-guided Visual Learning}
\label{sec:proprio_learning}

To enhance the robustness and accuracy of the policy-oriented visual representation, we introduce a proprioceptive supervision module. This module leverages the robot’s internal state signals—such as joint angles \( q_t \in \mathbb{R}^{n_q} \) and end-effector position \( x_t \in \mathbb{R}^3 \), where \( n_q \) is the number of degrees of freedom—as auxiliary supervision for the downstream visual projector.

Specifically, we design a decoder network \( D \) (the proprioception MLP) to predict the proprioceptive state from the final visual condition embedding. Following the VGGT-to-policy path above, this feature is the MLP output \(\mathcal{C}_t\) after Transformer encoding and average pooling:
\[
f_t = \mathcal{C}_t \in \mathbb{R}^{d_c},
\]
and the predicted proprioceptive state is given by:
\[
\hat{p}_t = D(f_t) \in \mathbb{R}^{n_q + 3},
\]
where the ground truth proprioceptive state is \( p_t = [q_t, x_t] \in \mathbb{R}^{n_q + 3} \).

The training objective includes a reconstruction loss on the proprioceptive state using the mean squared error:
\[
\mathcal{L}_{\text{proprio}} = \mathbb{E}_{t} \left[ \| p_t - \hat{p}_t \|^2 \right].
\]

This loss acts as an auxiliary signal during training to shape the downstream policy-oriented visual representation. Because the VGGT aggregator is frozen, gradients from \(\mathcal{L}_{\text{proprio}}\) update the trainable visual projector and decoder \(D\), but do not update VGGT itself; the resulting representation improves policy generalization in manipulation tasks.


\section{Experiments}
\label{sec:experiments}

\subsection{Benchmark Tasks and Evaluation Setup}

To systematically evaluate the effectiveness and generalization capability of our method, we select a representative subset of tasks from the MetaWorld benchmark~\cite{metaworld}. Specifically, we include the eight tasks on which the baseline methods—Diffusion Policy (DP) and its improved version DP3—achieve the lowest success rates, according to the appendix. These tasks are considered particularly challenging due to complex spatial reasoning, occlusions, or high-precision manipulation requirements. Additionally, we incorporate two relatively simple tasks (\textbf{Reach} and \textbf{Peg Unplug Side}) to assess the consistency of our method in standard settings.

\begin{figure*}
    \centering
    \includegraphics[width=0.99\linewidth]{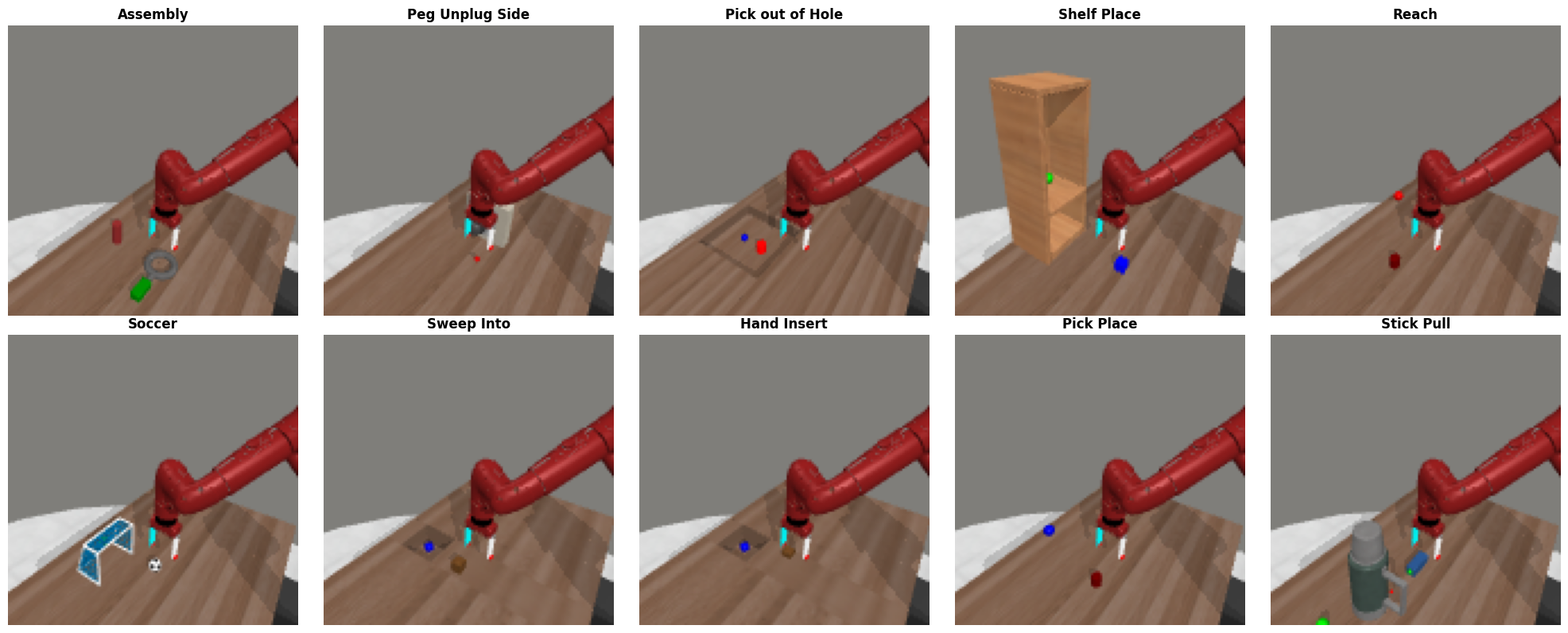}
    \caption{Visualization of ten tasks}
    \label{fig:tentasks}
\end{figure*}

The final evaluation set consists of the following ten tasks, as shown in Figure~\ref{fig:tentasks}: \textbf{Disassemble}, \textbf{Peg Unplug Side}, \textbf{Pick out of Hole}, \textbf{Shelf Place}, \textbf{Reach}, \textbf{Soccer}, \textbf{Sweep Into}, \textbf{Hand Insert}, \textbf{Pick Place}, and \textbf{Stick Pull}. All methods are evaluated using the standard MetaWorld protocol, with success rates computed over 20 test episodes per task. The training dataset consists of expert demonstrations, with 10 episodes, generated by a script policy. 

\subsection{Implementation Details}
The experiments are conducted using the PyTorch framework, with code adapted from the Diffusion Policy repository \citep{chi2023diffusionpolicy}. The random seed is fixed at 0 for reproducibility.

The pretrained VGGT aggregator is kept frozen throughout training. The trainable parameters comprise the downstream visual projector, the proprioception MLP, and the diffusion U-Net; consequently, auxiliary proprioceptive supervision does not update the VGGT aggregator.

\paragraph{Diffusion Policy Configuration}

The Diffusion Policy is implemented following the formulation in Section~\ref{sec:problem_formulation}, with specific architectural and training configurations detailed as follows. The model adopts a U-Net-1D architecture for the diffusion process. The encoder outputs a feature vector of dimension 64. The U-Net backbone uses a downsampling structure with channel dimensions and applies FiLM\cite{film} conditioning for integrating observation features.

For the diffusion process, we utilize the DDIM scheduler \citep{song2020denoising} with 100 training timesteps. During inference, we sample using 10 denoising steps. The model is trained with a prediction horizon of 16 steps. The input includes 2 observation steps, and observations are incorporated as global conditioning inputs. To enhance embodiment awareness, proprioceptive signals are also used as conditional inputs. 

\paragraph{Training and Evaluation Details}
We train all models for 3000 epochs using the AdamW optimizer~\citep{loshchilov2017adamw} with a learning rate of \(1.0 \times 10^{-4}\), \(\beta = [0.95, 0.999]\), \(\epsilon = 1.0 \times 10^{-8}\), and a weight decay of \(1.0 \times 10^{-6}\). A cosine learning rate scheduler is employed with 500 warmup steps. The training is conducted with a batch size of 128 using 8 workers.

To improve training stability and evaluation robustness, we apply an Exponential Moving Average (EMA) to the model parameters, with a decay rate configured by \(\gamma = 1.0\), power 0.75, and a maximum value of 0.9999. During training, the EMA-averaged policy is evaluated every 200 epochs by running 20 episodes per checkpoint. We report the final result as the best performance, computed as the average and standard deviation over the top five highest success rates across all evaluation checkpoints.


\begin{table}[t]
\centering
\small

\label{tab:metaworld-results}
\begin{tabular}{@{}lccc@{}}
\hline
\textbf{Task} & \textbf{DP} & \textbf{DP3} & \textbf{VGGT-DP} \\
\hline
Disassemble         & 43$\pm$7   & \textbf{69$\pm$4}   & 55$\pm$2.5 \\
Peg Unplug Side     & 74$\pm$3   & \textbf{75$\pm$5}   & 63$\pm$6 \\
Pick out of Hole    & 0$\pm$0    & 14$\pm$9   & \textbf{55$\pm$6} \\
Shelf Place         & 11$\pm$3   & \textbf{17$\pm$10}  & 10$\pm$0 \\
Reach               & 18$\pm$2   & 24$\pm$1   & \textbf{42$\pm$8} \\
Soccer              & 14$\pm$4   & 18$\pm$3   & \textbf{30$\pm$7} \\
Sweep Into          & 10$\pm$4   & 15$\pm$5   & \textbf{44$\pm$4} \\
Hand Insert         & 10$\pm$4   & 15$\pm$5   & \textbf{19$\pm$4} \\
Pick Place          & 0$\pm$0    & \textbf{12$\pm$4}   & 0$\pm$0 \\
Stick Pull          & 11$\pm$2   & 27$\pm$8   & \textbf{48$\pm$5} \\
\hline
\textbf{Average}    & 19.1       & 28.6       & \textbf{36.6} \\
\hline
\end{tabular}
\caption{
Success rates (\%) on 10 selected MetaWorld tasks. 
Results are reported as \textbf{mean ± standard deviation}.
Higher values indicate better task performance.
Bold numbers denote the best result for each task across all methods.
}
\end{table}

\subsection{Effectiveness on Task Success}

To better understand the performance of VGGT-DP across different tasks, we analyze its success rates in comparison with DP and DP3, focusing on three key observations:

\noindent\textbf{Strong Gains on Complex Spatial Tasks.}
VGGT-DP demonstrates substantial improvements on tasks that demand precise spatial reasoning and dynamic adaptability, such as \textbf{Pick Out of Hole}, \textbf{Sweep Into}, \textbf{Reach}, \textbf{Soccer}, and \textbf{Stick Pull}. These tasks benefit greatly from VGGT’s geometry-aware feature extraction and its ability to align visual and proprioceptive cues. The random token pruning module enhances robustness under partial observability. Furthermore, the fusion of proprioceptive signals strengthens closed-loop feedback, particularly in tasks requiring accurate end-effector alignment. Overall, these components synergistically improve policy generalization and trajectory stability in challenging settings.

\noindent\textbf{Limited Improvement on Simple Tasks.}
For relatively simple tasks such as \textbf{Peg Unplug Side} and \textbf{Disassemble}, VGGT-DP does not consistently outperform the baseline models. This can be attributed to two main factors: (i) these tasks typically require limited spatial reasoning and can be solved effectively using basic visual features, which DP and DP3 already capture well; (ii) the large-capacity VGGT encoder may introduce overparameterization and learning instability when the task complexity and data diversity are low. Moreover, the added components, such as token pruning and visual-proprioceptive alignment, may not provide significant benefits in such low-variance settings. They could even lead to marginal degradation due to optimization overhead.

\noindent\textbf{Failure Cases on Place Tasks.} 
Notably, VGGT-DP performs poorly on two \textbf{place} type tasks: \textbf{Shelf Place} and \textbf{Pick Place}, with success rates approaching zero. We hypothesize that this is due to the inherent challenges of these tasks, where the objects to be manipulated are often small, slender, or partially occluded. These characteristics increase the need for precise visual localization and grasp planning. While VGGT-DP utilizes multi-view spatial cues to enhance geometric reasoning, the limited visibility and fine-grained nature of the target objects may impede accurate perception and action prediction, ultimately resulting in failure to complete the placement.

\noindent\textbf{Summary.} VGGT-DP excels in high-complexity manipulation scenarios by leveraging its strong geometric representation and multimodal alignment capabilities. However, it may underperform on simple or poorly observed tasks due to overparameterization or incomplete visual context.

\begin{figure}[t]
    \centering
    \includegraphics[width=1\linewidth]{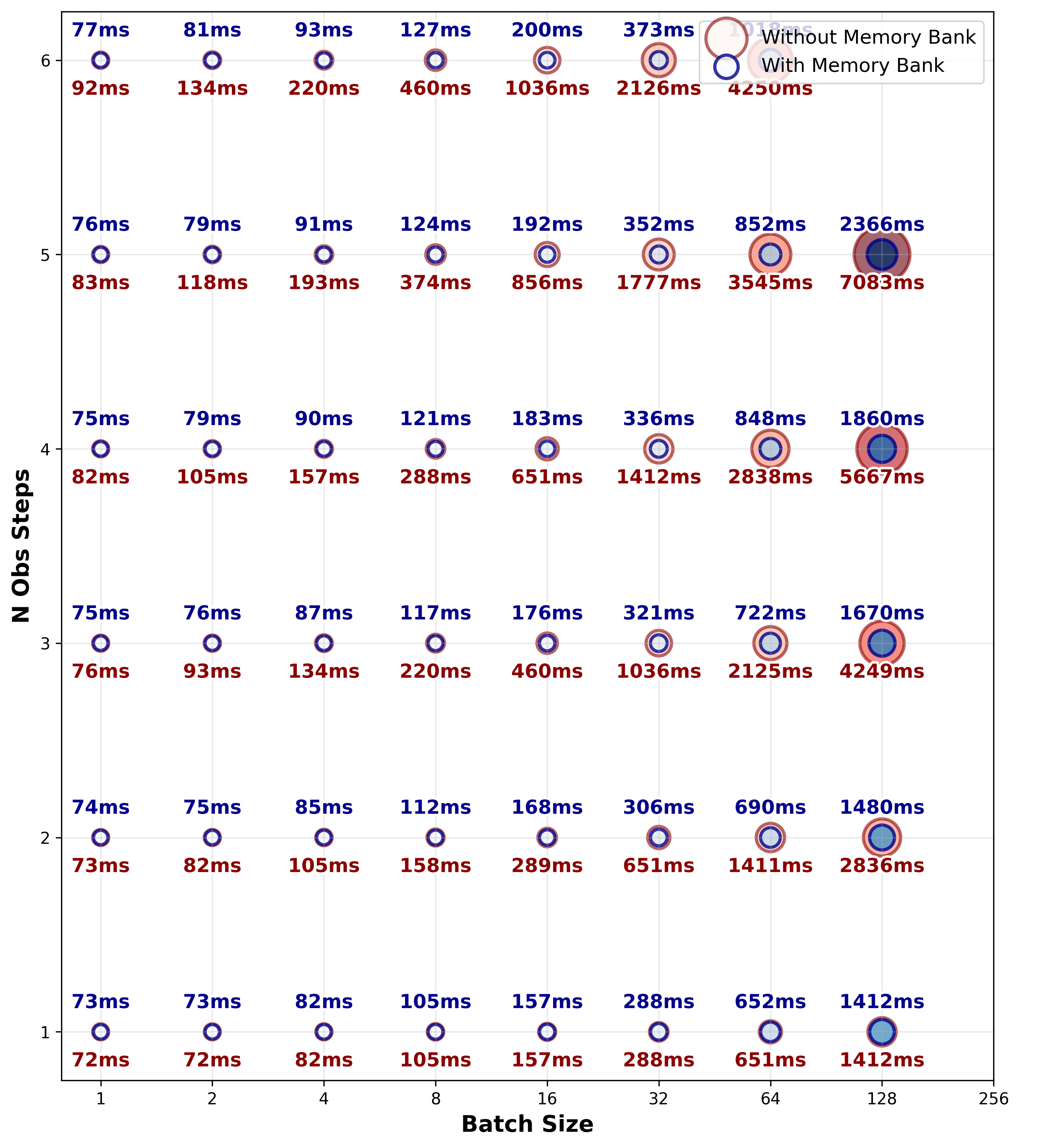}
    \caption{\textbf{Inference latency of the VGGT encoder under varying batch sizes and observation lengths.} Results with the Frame-Wise Token Reuse (FTR) mechanism are represented by blue circles, while those without FTR are shown in red. The FTR mechanism significantly reduces latency by reusing tokens across frames, particularly under larger temporal and batch dimensions.}
    \label{fig:inf_time}
\end{figure}

\paragraph{Efficiency of VGGT encoder}

We benchmark the VGGT encoder's inference latency under varying batch sizes and observation steps to assess the benefit of the Frame-Wise Token Reuse mechanism. For each configuration, we generate synthetic observations including RGB images, point clouds, and agent proprioception. Inference is conducted using a single GPU, synchronizing after each run to measure latency accurately. The number of observation steps \(T \in \{ 1,2,3,4,5,6\}\) and batch sizes \(\in \{1,2,4,8,16,32,64,128\}\) are systematically explored. As shown in Fig.\ref{fig:inf_time}, enabling the Frame-Wise Token Reuse mechanism leads to a significant reduction in inference time, particularly when the input temporal length and batch size are large. This is primarily because the FTR mechanism minimizes redundant computation of past observations. Although local inference is dominant in current robotics applications, our findings suggest strong potential for inference in cloud-based robot policy deployments.

\subsubsection{Viewpoint Perturbation Evaluation}
\label{sec:viewpoint_perturbation}

To evaluate the robustness of VGGT-DP under viewpoint variations, we conduct controlled perturbation experiments on the challenging \textbf{Stick Pull} task. This setting emulates realistic deployment scenarios where camera pose may deviate from the training distribution due to minor hardware shifts or dynamic environmental factors.

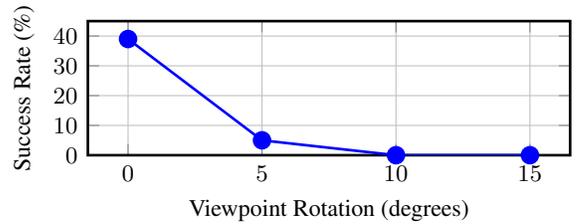
\begin{figure}[t]
    \centering
    \begin{tikzpicture}
        \begin{axis}[
            width=0.95\linewidth,
            height=0.4\linewidth,
            xlabel={Viewpoint Rotation (degrees)},
            ylabel={Success Rate (\%)},
            ymin=0, ymax=45,
            xtick={0,5,10,15},
            ytick={0,10,20,30,40},
            grid=major,
            mark size=3pt,
            line width=1pt,
            tick label style={font=\small},
            label style={font=\small},
        ]
        \addplot[
            color=blue,
            mark=*,
        ]
        coordinates {
            (0,39)
            (5,5)
            (10,0)
            (15,0)
        };
        \end{axis}
    \end{tikzpicture}
    \caption{\textbf{Viewpoint Robustness on StickPull Task.} Effect of viewpoint rotation on StickPull task success rate.}
    \label{fig:viewpoint}
\end{figure}

\paragraph{Experimental Setup.}
We simulate viewpoint perturbations by injecting random rotations into the camera pose at each observation step. Specifically, we apply noise to the roll, pitch, and yaw axes by independently sampling from a uniform distribution \([-{\delta}^\circ, +{\delta}^\circ]\), where \(\delta \in \{0, 5, 10, 15\}\). This ensures that for every timestep, the rendered camera view is subject to stochastic orientation noise, forcing the policy to operate under diverse visual perspectives. All other environmental factors and action spaces are kept unchanged to isolate the effect of viewpoint variation.

\paragraph{Results and Analysis.}
\ref{fig:viewpoint} presents the success rates of VGGT-DP on \textbf{Stick Pull} under increasing levels of viewpoint perturbation. At \(\delta=0^\circ\), the policy achieves a success rate of \(39\%\), but performance drops sharply to \(5\%\) at just \(5^\circ\), and completely fails (\(0\%\)) beyond \(10^\circ\). These results reveal that VGGT-DP, despite incorporating a large-scale pretrained visual encoder, lacks sufficient generalization to novel viewpoints.

This degradation suggests two potential limitations. First, the encoder's geometric priors, derived from 3D reconstruction tasks, may be overly reliant on consistent multi-view alignment, rendering the visual representation brittle under spatial shifts. Second, the training pipeline does not explicitly expose the policy to viewpoint variations, causing the learned features to overfit to the camera poses seen in demonstration data.

In summary, while VGGT-DP demonstrates strong performance in standard settings, its failure under mild viewpoint perturbations underscores the need for explicit view-robust training mechanisms, such as equivariant encoders or domain-randomized data augmentation, to ensure safe and reliable deployment in real-world robotic applications.

\section{Related Work}
\label{sec:related_work}

\subsection{Visual Imitation Learning in Robot Manipulation}

Visual motor policy has long been a central focus in the field of robot learning. Early end-to-end visual policy methods such as E2E-VMP \cite{levine2016end} and QT-Opt \cite{kalashnikov2018qt} demonstrated strong performance in manipulation tasks. More recently, imitation learning approaches such as the Diffusion Policy \cite{chi2023diffusionpolicy,ze20243d} and ACT \cite{zhao2023learningaloha} have shown impressive results across a wide range of manipulation scenarios.

Despite their success, it is widely recognized within the community that these methods struggle with limited generalization ability. To address this issue, many works have explored multimodal inputs, particularly language, to improve policy generalization and flexibility. Large-scale pretrained models such as RT series\cite{brohan2022rt1,zitkovich2023rt2}, OpenVLA\cite{kim2024openvla}, Pi series\cite{pi0,pi0.5} enable policies to understand natural language commands and execute semantically guided tasks. In these systems, diffusion-based or flow-matching models are often used as action decoders, due to their strong multimodal modeling capabilities and the high quality of the actions they generate, which enables more precise and coherent execution in complex manipulation tasks.

In addition, several works integrate multiple sensing modalities or spatial priors to enhance policy performance. For instance, the RDP series \cite{xue2025reactive} incorporates tactile sensing to address fine-grained tasks, while DP3 \cite{ze20243d} and SP-VLA \cite{qu2025spatialvla} leverage point cloud data to extract 3D structural information for more precise robot control. Recent methods such as Spatial Policy \cite{liu2025spatialpolicy} further highlight the importance of spatial-aware modeling for visuomotor manipulation. Other studies focus on improving the inference efficiency of policies, such as SP \cite{li2025sp} and block-wise diffusion \cite{ji2025block}, enabling real-time deployment. To further enhance viewpoint robustness and scene generalization, methods like EQP, SE3DP, and ET-SEED \cite{wang2024edp,tie2024etseed,zhu2025se} utilize equivariant representations over point clouds. While effective, these approaches often require high-fidelity 3D inputs, making them costly and hardware-intensive to deploy. In contrast, other methods, including RoboSplat \cite{yang2025robosplat} and DemoGen \cite{xue2025demogen}, adopt data augmentation strategies to improve robustness across diverse environments. To address the limitations of these methods, we introduce VGGT---a visual encoder pretrained on large-scale 3D reconstruction tasks---as the perception backbone in our policy framework. Experimental results demonstrate that this model significantly enhances performance across several challenging tasks, validating the potential of vision foundation models for robotic policy learning.

\subsection{Visual Foundation Model}

In recent years, pretrained vision foundation models (VFMs) have become an integral component in visual robot policies. Widely adopted general-purpose vision models, such as CLIP\cite{clip}, SigLIP\cite{zhai2023sigmoid}, DINOv2\cite{oquab2023dinov2}, JEPA\cite{assran2023selfjepa1,assran2025jepa2}, and R3M\cite{nair2022r3m}, leverage image-text alignment, self-supervised learning, or video-based pretraining techniques, thus exhibiting strong semantic representations and excellent cross-task transferability. These models have shown notable performance in image recognition, cross-modal semantic understanding, and video object analysis, particularly excelling within vision-language-aided (VLA) robot policy frameworks.

However, such general-purpose vision models primarily emphasize semantic perception and local feature extraction, lacking explicit understanding and consistent modeling of 3D spatial structures. This limitation becomes particularly prominent when robots are tasked with fine-grained manipulation, navigation planning, or real-time interactive tasks, significantly restricting the generalization capabilities of existing policies.

To overcome this challenge, recent studies have introduced visual foundation models specifically oriented towards 3D structure modeling, such as M3R\cite{cabon2025must3r}, D3R\cite{wang2024dust3r}, and VGGT\cite{wang2025vggt}. Notably, VGGT (Visual Geometry Grounded Transformer), pretrained on large-scale 3D reconstruction tasks, substantially enhances the model's implicit spatial understanding. VGGT predicts camera parameters, dense depth maps, 3D point maps, and 3D point tracks, providing coherent and structured spatial representations. Compared to general-purpose semantic models like CLIP, VGGT delivers superior support for robotic spatial reasoning and precise manipulation. Additionally, VGGT employs a purely feed-forward Transformer architecture, eliminating traditional multi-step optimization procedures, thus significantly improving efficiency and stability in real-world deployment.

Motivated by these advantages, we adopt VGGT as the perception backbone in robot visual policy models.

\section{Limitations}
\label{sec:limitation}

One key limitation of our proposed framework is the high computational overhead introduced by utilizing the VGGT encoder. Due to its large parameter count, the inference speed of our model is significantly constrained, limiting its practical applicability in real-time robotic manipulation scenarios. Future work will focus on exploring lighter yet effective encoding strategies to alleviate this computational bottleneck without substantially compromising the generalization performance.

\section{Conclusion}
\label{sec:conclusion}

In this work, we propose VGGT-DP, a biologically inspired visuomotor control framework that integrates a large-scale 3D-pretrained visual foundation model with a diffusion-based policy. Our design addresses two critical challenges in visual robotic manipulation: insufficient spatial representation and poor generalization. To this end, we introduce three key techniques: (i) a Frame-Wise Token Reuse mechanism that reduces redundant computation and improves inference efficiency; (ii) a random token pruning strategy that enhances robustness against partial observability; and (iii) a proprioception-guided visual learning module that aligns visual features with internal robot states, thereby improving feedback awareness and policy stability. Through extensive experiments on challenging MetaWorld tasks, we demonstrate that VGGT-DP outperforms strong baselines in tasks requiring precise control and long-horizon reasoning. However, the system's sensitivity to unseen viewpoints highlights the need for future improvements in viewpoint-invariant feature learning. Overall, our results confirm the promise of combining spatially grounded vision foundation models with probabilistic action decoders to build more generalizable and efficient robot control systems.
\bibliography{ref}
\end{document}